# Three-dimensional Optical Coherence Tomography Image Denoising through Multi-input Fully-Convolutional Networks

Ashkan Abbasi[a], Amirhassan Monadjemi[a,*], Leyuan Fang[b,*], Hossein Rabbani[c], and Yi Zhang[d].

[a] *Artificial Intelligence Department, Faculty of Computer Engineering, University of Isfahan, Isfahan, Iran. (E-mail: monadjemi@eng.ui.ac.ir)*

[b] *College of Electrical and Information Engineering, Hunan University, Changsha, China. (E-mail: fangleyuan@gmail.com)*

[c] *Department of Biomedical Engineering, Medical Image and Signal Processing Research Center, School of Advanced Technologies in Medicine, Isfahan University of Medical Sciences, Isfahan, Iran.*

[d] *College of Computer Science, Sichuan University, Chengdu, China*

*Asterisk indicates the corresponding author.*

*Abstract*—**In recent years, there has been a growing interest in applying convolutional neural networks (CNNs) to low-level vision tasks such as denoising and super-resolution. Due to the coherent nature of the image formation process, optical coherence tomography (OCT) images are inevitably affected by noise. This paper proposes a new method named the multi-input fully-convolutional networks (MIFCN) for denoising of OCT images. In contrast to recently proposed natural image denoising CNNs, the proposed architecture allows the exploitation of high degrees of correlation and complementary information among neighboring OCT images through pixel by pixel fusion of multiple FCNs. The parameters of the proposed multi-input architecture are learned by considering the consistency between the overall output and the contribution of each input image. The proposed MIFCN method is compared with the state-of-the-art denoising methods adopted on OCT images of normal and age-related macular degeneration eyes in a quantitative and qualitative manner.**

1. INTRODUCTION

Optical coherence tomography (OCT) is a noninvasive imaging modality which is widely applied in the diagnosis and treatment planning of various ocular diseases [1]. Due to interferometry nature of the image formation process, noise corruption is inevitable during OCT imaging. The presence of noise heavily degrades the image quality and complicates the image analysis. Quality of OCT imaging could be improved either by applying higher incident power or longer exposure time [2]. Both of these options cannot be used because 1) The incident power is limited by the safety guidelines, and 2) The imaging speed is an important factor to avoid motion artifacts caused by the fixation eye movements [3] or enable 3-D volumetric imaging. Thus, image denoising is an essential step in many OCT image analysis tasks.

There exist many proposed methods regarding the OCT image denoising, where, the early spatial filtering approaches [4] are based on computing local statistics of the degraded image in the spatial domain. Image content can be transformed into another domain like the filtering response domain [5] or multi-resolution domain [6,7], where image statistics are modeled more efficiently. Although the results obtained through the transform domain approaches are promising, their limited modeling ability generally result in smoothing or unexpected artifacts [8].

The image modeling ability has been improved by the introduction of patch-based approaches, because patches have lower dimensions compared to the whole image, and are easier to model. Moreover, patches capture local image statistics, thus edges and local structures are treated better. Some of the most successful modeling approaches consist of the Markov random field (MRF) [9], sparse representation [10], and Gaussian mixture models (GMM) [11]. Some of the recent studies where the patch-based sparse representation is applied to OCT image reconstruction consist of [12–16]. Recently, a variant of GMM [17] is applied to OCT image denoising with promising results [18]. Most of the mentioned approaches are enhanced greatly by applying the nonlocal similarity [19] in natural images. Although in this realm reasonable success is recorded due to the mentioned approaches, they mostly rely on computationally expensive optimization algorithms in the reconstruction stage. Moreover, patch aggregation through averaging negatively affects the effectiveness of the image model [10].

Deep learning approaches are proven to be highly effective in many high-level vision tasks [20,21] and are successfully applied in medical image recognition tasks, including classification detection and segmentation. The great success of the neural networks and the progress made in their training methods pave the way for applying the neural networks as a promising alternative approach to deal with the image denoising problems. The authors in [22] are the first who revealed that denoising through a convolutional neural network (CNN) could outperform several well-known methods. In [23], it is revealed that a multi-layer perceptron (MLP) can have comparable performance to the benchmark BM3D [24]. The MLP has a fully connected architecture which makes the computation of both the training and inference intensive. Therefore, considerable attention has been recently given to CNNs [25,26].

In this paper, following [12–15] where they learned mappings from noisy images to high signal-to-noise (SNR) images using sparse representations, a specifically designed CNN is proposed for this purpose. To the best of the authors' knowledge, the proposed method is the first CNN-based OCT image denoising method. With the advances made in OCT imaging, the acquisition of 3-D volumetric scans of the retina is widely applied in its clinical sense.

Practical application of the information from nearby images is a promising manner in reducing noise [14,16]. Although the neural network-based denoising approaches are efficient, the focus of most is on 2-D gray-scale image denoising [27]. The issue of how to effectively apply the high correlations among nearby OCT images to reduce noise by a CNN is not assessed yet.

The objective here is to develop a network architecture for OCT image denoising where the high correlations are effectively used among nearby OCT images through a multi-branch network. Each branch is a fully-convolutional network (FCN) with the objective to reduce the noise of its input. This method is named the multi-input FCN (MIFCN). The results obtained through these FCNs are fused by an intermediate weighted averaging module inspired by the nonlocal mean weighting mechanism [19], which produces weight matrices for each branch. Then, Hadamard products of the weight matrices and the outputs of each branch are computed to generate the averaging module's output. The output is processed by another set of convolution layers to generate the final reconstructed image. Because the weighting mechanism suppresses the useless contributions among nearby images, it can be assured that the proposed method can capture their correlations while it is insensitive to small variations in the inputs. The parameters of the proposed MIFCN method can be learned by optimizing a loss function that is specifically designed to enable end-to-end training of the overall architecture.

The rest of this paper is organized as follows. In the following section, we briefly review related works. Next, we describe the proposed MIFCN method in Sec. 3. Training the parameters are presented in Sec.4. The experimental results are presented in Sec.5, and the paper is conducted in Sec. 6.

2. RELATED WORKS

*2.1* CONVOLUTIONAL NEURAL NETWORK

CNN is a multilayer architecture with an input layer, an output layer, and multiple hidden layers. The hidden layers mostly consist of convolutional and pooling layers. The last hidden layers can be fully connected layers for global decision-making. The convolution and pooling layers enable the whole structure to extract a hierarchical representation of the data, where the shallower layers concentrate on low-level features, while the deeper layers represent higher-level features [28]. Each layer is composed of multiple feature maps. Each unit in a feature map (neuron) is computed by a local operation (i.e., convolution or pooling), on the previous layer. By contrast, in a fully connected layer, each neuron is connected to every neuron in the antecedent layer. The local connectivity greatly reduces the number of parameters to be learned and captures natural images' local statistics. Moreover, these local operations can be run on arbitrary-sized inputs. Therefore, Long et al. [29] proposed a variant of CNNs by casting fully connected layers into convolutions and named it FCN, which makes the FCN a natural choice for image transformation tasks.

*2.2* NETWORK ARCHITECTURES

In general, the main purpose of successive convolution and pooling layers in CNNs is hierarchical feature extraction. However, the pooling or sub-sampling, in any form, leads to a loss of spatial information [30,31].

Therefore, a network architecture without subsampling layers is chosen as a building block for implementing our proposed MIFCN method.

The context aggregation network (CAN) is recently proposed for semantic segmentation [30], and it consists mainly of convolution layers with dilated convolutions. The dilated convolution provides the means to aggregate the contextual information without the need for any form of subsampling. More recently, a fast and compact variant of CAN is applied for approximating some image processing operators [31].

## 2.3 CONVOLUTION LAYER AND DILATED CONVOLUTION

Convolution layer is the main ingredient of a CNN architecture [29]. Each convolution layer is composed of several feature maps of the same size, where each feature map highlights the regions in the input that are most similar to its corresponding filter. These filters are learned in the sense that they eventually activate suitable features for a given task. During the forward pass, the feature maps are computed from the previous layer through Eq. (1):

$$F_i^l = b_i^l + \sum_j F_j^{l-1} *_d K_{i,j}^l, \qquad (1)$$

where, $F_i^l$ is the $i^{\text{th}}$ feature map of layer $l$, $F_j^{l-1}$ is $j^{\text{th}}$ feature map of the antecedent layer $(l-1)$, $K_{i,j}^l$ is a convolution kernel, the operator $*_d$ is the dilated convolution with dilation $d$, and $b_i^l$ is a bias. By increasing the dilation $d$, the filter can tap locations separated by the factor $d$ without losing resolution [30]. In its mathematical sense, the $d$-dilated convolution at location $x$ between a feature map and a kernel is expressed through Eq. (2):

$$(F *_d K)(x) = \sum_{a+db=x} F(a)K(b), \qquad (2)$$

## 2.4 ACTIVATION LAYER

To enhance the representation ability of neural networks, the results of a convolution layer usually pass through a point-wise nonlinearity. A favorite activation function is the rectified linear function (ReLU), and because its outputs are zero for all negative values, some neurons may die (dying ReLU problem). The leaky variants of ReLU are applied to overcome this problem. Here, the leaky ReLU (LReLU) [32] defined as $\sigma(x) = \max(\alpha x, x)$ is applied where the constant parameter $\alpha$ (named the leak parameter) determines the slope for the negative values.

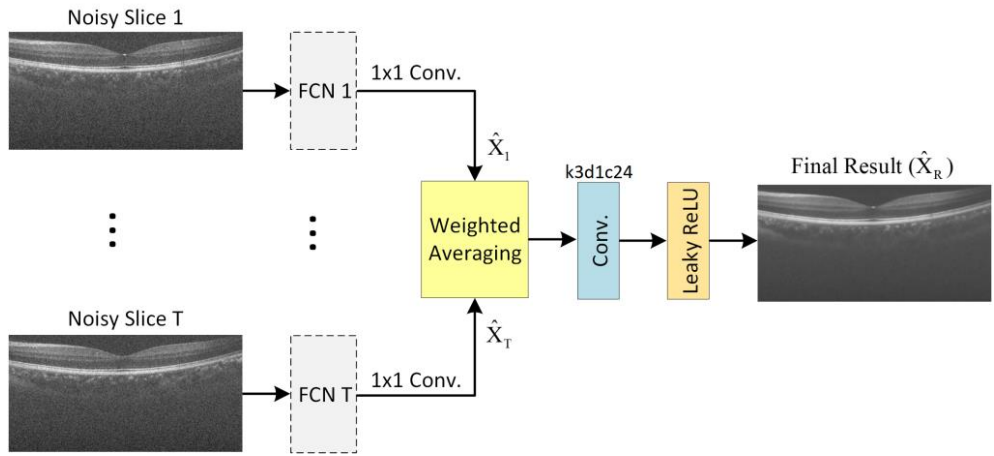

Fig. 1. The overall structure of the proposed MIFCN method for OCT image denoising. The numbers in front of the letters *k*,

*d*, and *c* represent kernel size, dilation rate, and some feature maps, respectively.

3. THE PROPOSED MIFCN METHOD FOR OCT IMAGE DENOISING

Given an OCT image observation $Y_1$ (main image) with *T-1* number of its nearby OCT images, it is sought to design a network that can effectively utilize the correlations among these *T* inputs $\{Y_1, ..., Y_T\}$ and reduce noise of the main image in an effective manner. This problem can be considered as a regression problem [12,13]. Here a multi-input architecture is proposed that can be learned in an end-to-end manner. The overall architecture of the proposed MIFCN method is shown in Fig. 1. Each branch consists mainly of convolution layers with the objective to reduce the amount of noise in its input, thus, making each branch an FCN. Here, a weighted averaging module is applied to combine the results based on their similarity to the main image ($Y_1$). This architecture is followed by another convolution and activation layers to enhance its modeling ability and yield better results.

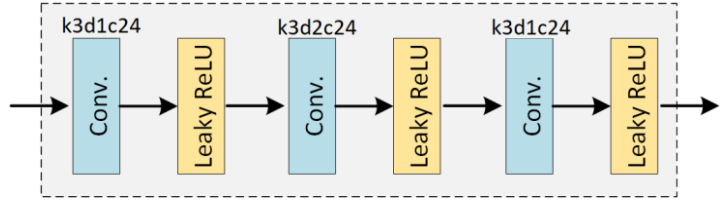

Fig. 2. The structure of each FCN applied in the overall architecture here (Fig. 1) of the proposed MIFCN method. The numbers in front of the letters *k*, *d*, and *c* represent kernel size, dilation rate, and number of feature maps, respectively.

The network for each branch is designed according to Fig. 2. A similar network structure is applied to all branches. This structure consists of three convolution layers, all with the same number of feature maps *C* = 24. The dilation rate (*d*) for each one of the hidden layers are set to 1, 2, and 1, respectively. The output of each convolution layer is passed through an activation layer. Let the input to the $t^{th}$ branch be $F^0 = Y_t \in R^{M \times N}$, where $t \in \{1,2, ..., T\}$, then, each feature map in the convolution layers is computed through Eq. (3):

$$F_i^l = \sigma(b_i^l + \sum_{j=1}^{C} F_i^{l-1} *_d K_{i,j}^l), \qquad (3)$$

where, $l \in \{1,2,3\}$ is the layer number, and $K_{i,j}^l$ is a $3 \times 3$ kernel. Then, another convolution layer without activation layer is applied for reconstructing an output image at the end of each branch:

$$F^4 = b_i^4 + \sum_{j=1}^{C} F_i^3 *_1 K_{i,j}^4 = \hat{X}_t, \qquad (4)$$

where, $K_{i,j}^4$ is a $1 \times 1$ convolution kernel.

During the forward pass, the outputs of branches result in a set of noise reduced images $\{\hat{X}_1, ..., \hat{X}_T\}$. Because the inputs are nearby images, the outputs of branches have some spatial correlations. However, there might be slight variations among these images. Thus, simple pixel by pixel averaging can result in blurring artifacts [33]. Motion compensation algorithms can be used with the expense of high computational cost and tolerating error (even for noise-free images) [34]. Here, inspired by the nonlocal mean (NLM) weighting mechanism [19], which is almost robust to slight variations in patches [33,35], we propose a module that compares each pixel of the main branch output ($\hat{X}_1$) to the corresponding pixels from the other outputs $\{\hat{X}_2, ..., \hat{X}_T\}$. The proposed module assigns weights to pixels of denoised nearby images. Each pixel in the output of the proposed module is simply formed by a weighted combination of a pixel from the main branch output and the corresponding pixels from the other outputs; consequently, the output of the averaging module is computed through Eq. (5):

$$\bar{X} = \sum_{t=1}^{T} \hat{X}_t \, o P_t, \qquad (5)$$

where, $\bar{X}$ is the output of the weighted averaging module, $\hat{X}_t$ is the output at the end of each branch, $P_t$ is the weight matrix with the same size as $\hat{X}_t$, and the operator $o$ is the Hadamard product. Each entry in the weight matrix ($P_t$) reflects the similarity among the corresponding pixels in the main branch output ($\hat{X}_1$) and the $t^{th}$ branch output ($\hat{X}_t$), therefore, the maximum scores are always assigned to the main branch pixels where the entries of $P_1$ are bigger than the others. To compute the weight matrix, first, it is necessary to compute the differences among the corresponding pixels from denoised nearby images through Eq. (6):

$$D_t = (\hat{X}_1 - \hat{X}_t)^2, \; t \in \{1,2, ..., T\}, \qquad (6)$$

where, $D_t$ is the intensity difference matrix, the elements of which are the differences among the pixels from the main branch ($\hat{X}_1$) and pixels from the $t^{th}$ branch ($\hat{X}_t$). This matrix can be applied in computing the exponentially decaying weights for pixels of an image through Eq. (7):

$$W_t = \exp(-D_t \backslash h), \qquad (7)$$

where, $W_t$ is the matrix of weights associated to each pixel in the $t^{th}$ branch prediction, and $h$ is a constant parameter. To assure that these weights sum to one, the weight matrix is normalized using Eq. (8):

$$P_t = W_t \backslash (\sum_{t=1}^{T} W_t), \qquad (8)$$

The result of the weighted averaging module is processed through another set of convolution layers, Fig. 1, where the output of the averaging module ($\bar{X}$) is fed into a convolution followed by LReLU activation layers to obtain the last hidden feature map ($F_i^5$). This feature map ($F_i^5$) is converted to the final image through a 1x1 convolution layer as follows:

$$F^6 = b_i^6 + \sum_{j=1}^{C} F_j^5 *_1 K_{i,j}^6 = \hat{X}_R, \qquad (9)$$

where, $\hat{X}_R$ denotes the final result of the proposed 3-D reconstruction method.

Before concluding this section, it is worth mentioning that unlike the original NLM weighting mechanism [19,33] which is based on comparing small patches around each pixel, here, the weights are computed based on comparing pixels between nearby denoised images. We have experimentally found that this pixel by pixel averaging module is strong enough to provide plausible results and avoid extra computations. The reasons can be explained as follows: 1) noise is expected to be reduced in each branch, thus pixel by pixel comparison is more robust compared to such a comparison among clear noisy images, and 2) this module can be compared with local filtering approaches [4]. In local filtering approaches, a pixel is denoised based on the weighted average of pixels surrounding it. Here, instead of considering a neighborhood around each pixel, the corresponding pixels from the nearby images are used. Therefore, it is reasonable to apply the same principles.

## 4. LEARNING THE PARAMETERS

Because FCN is not sensitive to the input size, an FCN can be trained for denoising by using only patches. In fact, training an FCN for denoising using patches have empirically shown to be helpful [25,26]. This is because it enables the network to capture more local information. As shown in Fig. 1, the proposed MIFCN architecture has $T$ branches. To train the parameters of all branches in a simultaneous manner, $T$ patch pairs are needed. In contrast to the test dataset, the training dataset does not include nearby OCT images. Because the training procedure is based on using patches, we can find similar patches for each patch. By following a simple procedure, we can create a dataset of patches and the similar ones as a training dataset. Here, first, $N$ patches of size $p_1 \times p_2$ pixels are extracted from all high SNR images and next, for each patch in an image, the $T$ most similar patches (including the patch itself) are collected by nonlocal searching [19] in that image. By extracting the corresponding noisy patches for each patch, the following training set is created:

$$D = \left\{\left\{\left(y_1^{(j)}, x_1^{(j)}\right), \ldots, \left(y_T^{(j)}, x_T^{(j)}\right)\right\}\right\}_{j=1}^{N}, \quad (10)$$

where, $y_t^{(j)}$ is the $t^{th}$ similar patch for the j$^{th}$ noisy patch and $x_t^{(j)}$ is its corresponding high SNR patch.

Given the training dataset D, a loss function is needed to train the set of parameters $\theta$ of the network architecture, which $\theta$ includes kernels and biases of all feature maps. The widely applied mean squared error (MSE) is the common choice for image reconstruction purposes. Nevertheless, here, the proposed MIFCN architecture cannot be trained using pure MSE. In the MSE, only the error between predicted outputs and desired outputs are of concern. Because the proposed MIFCN architecture has multiple branches, training its parameters using only MSE might result in a useless branch with zeros as outputs. Instead, a loss function is designed where the consistency between the overall output and the contribution of each branch are taken into account. This loss function enables training the architecture in an end-to-end manner:

$$J(\theta) = \frac{1}{N}\sum_{i=1}^{N}\sum_{t=1}^{T}\left(\hat{X}_t^{(i)} - X_t^{(i)}\right)^2 + \frac{1}{N}\sum_{i=1}^{N}\left(\hat{X}_1^{(i)} - \hat{X}_R^{(i)}\right)^2, \quad (11)$$

where, $N$ is the number of training patches, $T$ is the number of branches, $\hat{X}_t^{(i)}$ is the output of the last feature map in the $t^{th}$ branch for the i$^{th}$ noisy training patch, $\hat{X}_R^{(i)}$ is the final output of the architecture for the i$^{th}$ noisy training patch, and $X_t^{(i)}$ is the corresponding high SNR patch.

In Equation (11), the first term encourages the similarity between the result of each branch and the corresponding high SNR patch. This is because the last feature map should be a noise reduced version of its input. The second term encourages the similarity between the final output and the prediction of the first branch (or main branch). In this way, we can ensure that slight variations in the inputs cannot negatively affect the final output.

For training, the loss function $J(\theta)$ is minimized using a gradient descent based optimizer [36]. The network is trained with the augmented data generated by horizontal and vertical flipping, and +90-degree rotation. All of the training data are presented for 60 epochs; in the first 30 epochs, the learning rate is set to 0.0001, and next, it is set to 0.00001 for the remaining epochs. The total training time for this architecture is less than two hours.

## 5. EXPERIMENTAL RESULTS

In this section, we present experimental results of the proposed MIFCN method. We compare the proposed MIFCN methods with some of the well-known state-of-the-art denoising methods. The source code of our method will be made publicly available on the website (https://github.com/ashkan-abbasi66/MIFCN). Also, all of the visual results of the proposed method and compared methods are now available on the website.

### 5.1 THE DATASETS

To train and evaluate the proposed MIFCN method, we have used the spectral domain OCT (SDOCT) datasets that were made publicly available by [12–14]. All of these images are captured by a Bioptigen SDOCT imaging (Durham, NC, USA) from 28 subjects with normal and age-related macular degeneration (AMD) eyes. In the training part, there exist ten pairs of noisy and high SNR images. The high SNR images were acquired by registration of azimuthally repeated OCT images from the fovea [12–14]. The rest of the images are used as a test dataset. For each test image, four noisy nearby OCT images are also provided together with a high SNR image, consequently, an image denoising algorithm can exploit one or more OCT images to reconstruct a high-quality OCT image.

### 5.2 THE QUANTITATIVE METRICS

The performance of the proposed MIFCN method is assessed by different image reconstruction metrics. The peak-signal-to-noise-ratio (PSNR), mean-to-standard-deviation ratio (MSR) [37], contrast-to-noise-ration (CNR) [38], and equivalent number of looks (ENL) [6] are used here. Due to the availability of high SNR images, PSNR is computed as a widely accepted metric in this scenario. This metric is defined based on the intensity differences between the output and a reference image. The other metrics do not need the reference images and are computed locally. Therefore, a few regions of interest (ROIs) are selected from the images. The contrast between foreground regions (e.g., red box #2-#6 in Fig. 3) and background noise is measured through the CNR metric. The background noise is computed in the background region (e.g., red box #1 in Fig. 3). The CNR metric is big when ROIs contain prominent features. The MSR is a sign of good feature recovery without considering the background regions. Smoothness in background regions is assessed by ENL. Large ENL values indicate a stronger noise is smoothing in background areas [6], moreover, the Wilcoxon signed-rank test is applied to show the statistical differences between the proposed MIFCN method and the compared methods.

### 5.3 THE COMPARED METHODS

The proposed MIFCN method is compared with some of the well-known denoising methods. The comparison methods consist of: K-SVD denoising algorithm [10], block matching and 3-D filtering (BM3D) [24], spatially adaptive iterative singular-value thresholding (SAIST) [39], patch group based Gaussian mixture model (PG-GMM) [40], block matching and 4-D filtering (BM4D) [41], and segmentation based sparse reconstruction (SSR) [14].

The K-SVD denoising algorithm [10] is a well-known sparse representation based image denoising method, where, the sparse representation over a learned dictionary is applied to remove noise. The benchmark BM3D [24] combines the advantages of sparsity-based image modeling and nonlocal similarity within each group of similar patches. The BM4D [41] is an extension of BM3D for the volumetric data. In BM4D, groups of similar cubes are collaboratively filtered to reduce noise, therefore, the BM4D can naturally capture the correlation among multiple images. A low-rank approach is applied in SAIST [39] to characterize the local and nonlocal variations in a group of similar patches. In contrast to the existing nonlocal image restoration methods which are based on nonlocal similarity of corrupted patches, the PG-GMM [40] learns a nonlocal prior from an external training dataset consists of groups of similar patches. The SSR [14] is a recently proposed OCT image reconstruction algorithm which performs sparse representation over learned dictionaries for each layer. The dictionary for each layer is learned/selected using a segmentation algorithm. Thus, the SSR combines a good modeling approach (i.e., sparse representation over learned dictionaries) and a good model selection strategy.

*5.4* THE ALGORITHM PARAMETERS

Most of the parameters of the proposed MIFCN method, including kernels and biases of feature maps, are learned automatically from the training data. The constant parameter $\alpha$ of LReLU function was set to 0.2 and the identity initialization method [30] is adopted to initialize the kernels and biases of feature maps. The number of branches ($T$) is, in general, dictated by the imaging configuration (more specifically, it is based on the azimuthal resolution of the OCT volume). This is because the contributions of images with large difference in contents are to be avoided here. However, there is a constant parameter $h$ in the exponential weight function (7) that can be used to control the amount of contributions from nearby OCT images. The test dataset here has five images per subject. Therefore, the number of input branches $T$ is set to 5. The constant parameter $h$ is experimentally set to 400. Setting a smaller value for $h$ weakens the contributions of nearby images.

For training, ten pairs of noisy and high SNR images are used. The patches of size 15×15 pixels are extracted. Bigger size patches may cause blurring artifacts because of losing small details and patches of small size are more likely to capture noise [26]. Because there exists a large background portion in OCT images, first, a portion containing the retina from each training image is manually cropped and next, patches are extracted with as less overlap as possible to increase the variety of training samples. A total of 400 patch pairs are extracted from each training image pair. Flipping and rotation are applied to augment the training data by a factor of 3. Therefore, the number of training samples ($N$) in Equation (10) is $400 \times 10 \times 3$. After training the model, all of the mentioned parameters are kept unchanged during the experiments. The parameters of the compared methods are optimally assigned or set according to their original papers.

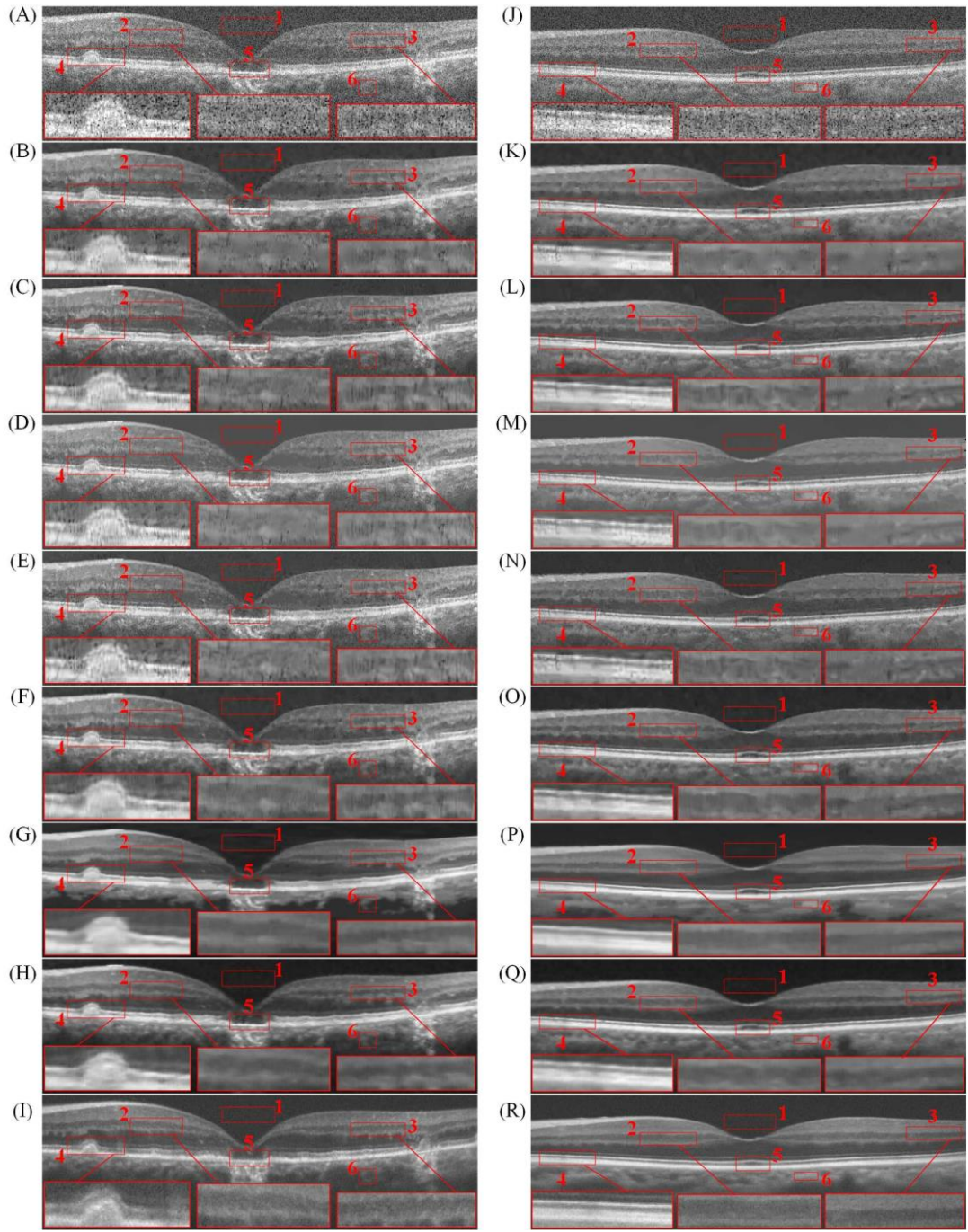

Fig. 3. Comparison of two denoised images by the compared methods: first column: (A) Original Noisy Image, (B) KSVD Denoising (PSNR = 26.05), (C) BM3D (PSNR = 26.25), (D) SAIST (PSNR = 26.01), (E) PG-GMM (PSNR = 26.1), (F) BM4D (PSNR = 26.48), (G) SSR (PSNR = 26.89), (H) proposed MIFCN method (PSNR = 27.49), (I) registered and averaged images and second column: (J) Original Noisy Image, (K) KSVD Denoising (PSNR = 26.13), (L) BM3D (PSNR = 26.02), (M) SAIST (PSNR = 26.16), (N) PG-GMM (PSNR = 25.89), (O) BM4D (PSNR = 26.54), (P) SSR (PSNR = 27.06), (Q) proposed MIFCN method (PSNR = 27.56), (R) registered and averaged images.

## 5.5 RESULTS FOR OCT IMAGE DENOISING

The two denoising results of the proposed MIFCN method are shown in Fig. (3), where, the corresponding high SNR images are shown at the bottom of each column. As can be seen, K-SVD, BM3D, SAIST, and PG-GMM attenuate the noise while result in apparent visual artifacts. Exploiting self-similarity in BM3D, SAIST, and PG-GMM allows better reconstruction of retinal layers. The SAIST method is based on a low-rank approach which is a powerful technique, especially for background regions. However, this method results in more artifacts especially in retinal layers compared to other methods (e.g., compare red box#4 in Fig. 3, (D) and (M) with their corresponding boxes in other images). The visual results confirm that BM4D, SSR, and the proposed method can better preserve layer structure due to exploiting correlation among nearby OCT images. Similar to the results of BM3D (e.g., Fig. 3, (C) and (L)), applying fixed bases in BM4D limits modeling ability and results in the formation of visible artifacts. However, there are fewer artifacts in the results of BM4D compared to the BM3D's results. This might be attributed to finding better matches by grouping 3-D patches in BM4D instead of grouping 2-D patches in BM3D. Although the SSR applies learned dictionaries for reconstruction of each layer, its reconstruction results are too smooth. The layer boundaries are reconstructed very well, but the smoothness can be easily observed from the background regions (vitreous and sclera) and cloudy appearance choroidal region below retinal layers.

The visual results can be validated by the average quantitative results, tabulated in Table I and II. In Table I, the three quantitative metrics (i.e., MSR, CNR, and ENL) which are widely applied in evaluating OCT reconstruction algorithms [6,14] are reported. The average PSNR results are tabulated in Table II. These quantitative results reveal that the proposed MIFCN method performs reasonably well in terms of all metrics, except for the mean of the ENL values. This is because ENL measures smoothness in background regions, therefore, these high ENL values for SAIST are attributed to the strong noise suppression in background areas due to exploiting a low-rank strategy. For the SSR method, smoothing textures is the main cause of yielding high ENL values. However, because SSR can well preserve the layers' structures, other metrics have relatively high values. All of these qualitative and quantitative results suggest that the proposed MIFCN method outperforms the other methods for OCT image denoising.

Table I

Mean and standard deviation (SD) of the MSR, CNR, and ENL results for denoising 18 foveal images by the compared methods. Where p<0.05, the metrics for each test method are considered statistically significant and were marked by "*". Best results in the mean values are shown in bold.

| Method | MSR | | | CNR | | | ENL | | |
|---|---|---|---|---|---|---|---|---|---|
| | Mean | SD | $p$ value | Mean | SD | $p$ value | Mean | SD | $p$ value |
| K-SVD | 7.53 | 1.26 | 8.91E-04* | 3.39 | 0.52 | 4.80E-07* | 776.33 | 150.54 | 5.61E-14* |
| BM3D | 6.86 | 0.96 | 1.34E-08* | 3.21 | 0.45 | 2.33E-10* | 1228.17 | 511.56 | 3.60E-10* |
| SAIST | 7.52 | 1.45 | 1.72E-03* | 3.19 | 0.47 | 4.84E-10* | **5752.29** | 1142.22 | 4.48E-10* |
| PG-GMM | 7.17 | 1.20 | 8.64E-06* | 3.22 | 0.48 | 1.09E-09* | 995.42 | 314.31 | 5.21E-13* |
| BM4D | 7.07 | 0.81 | 1.67E-09* | 3.31 | 0.44 | 4.73E-10* | 1037.98 | 262.66 | 4.02E-14* |
| SSR | 8.04 | 0.92 | 1.40E-03* | 3.57 | 0.49 | 5.74E-06* | 5225.34 | 3236.76 | 5.57E-03* |
| MIFCN | **8.38** | 0.94 | | **3.75** | 0.52 | | 2750.75 | 400.98 | |

Table II

Mean and standard deviation (SD) of the PSNR (dB) results for denoising 18 foveal images by the compared methods. Where p<0.05, the metrics for each test method are considered statistically significant and were marked by "*". Best results in the mean values are shown in bold.

| Method | PSNR | | |
|---|---|---|---|
| | Mean | SD | $p$ value |
| K-SVD | 26.21 | 2.68 | 1.54E-06* |
| BM3D | 26.18 | 2.68 | 1.60E-07* |
| SAIST | 26.15 | 2.73 | 1.30E-06* |
| PG-GMM | 26.08 | 2.67 | 6.96E-07* |
| BM4D | 26.66 | 2.74 | 2.41E-07* |
| SSR | 27.23 | 2.86 | 9.48E-01 |
| MIFCN | **27.37** | 2.73 | |

The average run-time (in seconds) for denoising obtained using the compared methods are tabulated in Table III. All experiments are run on a desktop PC with an Intel® i7-7700K CPU at 4.2 GHz, 16 GB of RAM, and a GPU of NVIDIA GeForce GTX 1080 Ti. According to this table, the proposed MIFCN method has the least run-time both on CPU and GPU.

Table III

Average Runtime (seconds) for denoising 18 foveal images by the compared methods. Best result is shown in bold.

| Method | Runtime (seconds) | |
|---|---|---|
| K-SVD | | 28.61 |
| BM3D | | 6.89 |
| SAIST | | 69.30 |
| PG-GMM | | 52.72 |
| BM4D | | 46.88 |
| SSR | | 16.31 |
| MIFCN | CPU: | 1.008 |
| | GPU: | **0.064** |

*5.6 Effects of Different Values of The Parameter $h$*

The objective here is to reduce the noise of an OCT image by exploiting the correlation among nearby OCT images. The number of contributions from other branches (or nearby images) controlled through a constant $h$ in Eq. (7). A few visual results are presented in Fig. 4, where, if a very small value is selected for $h$ (Fig. 4 (C)), the final result ($\hat{X}_R$) is almost identical to the prediction of the main branch ($\hat{X}_1$), because very small weights are assigned to each pixel of other branches. As the value of this constant increases, the amount of contributions from other denoised nearby images increase in a continuous manner. Comparing Fig. 4 (D) and (F) clearly indicates that in Fig. 4 (D) small features become blurry and the reconstruction result suffers from artifacts. Because the constant value of *h* has a significant effect on the final result, the effects of different values for this parameter are assessed in an experimental manner.

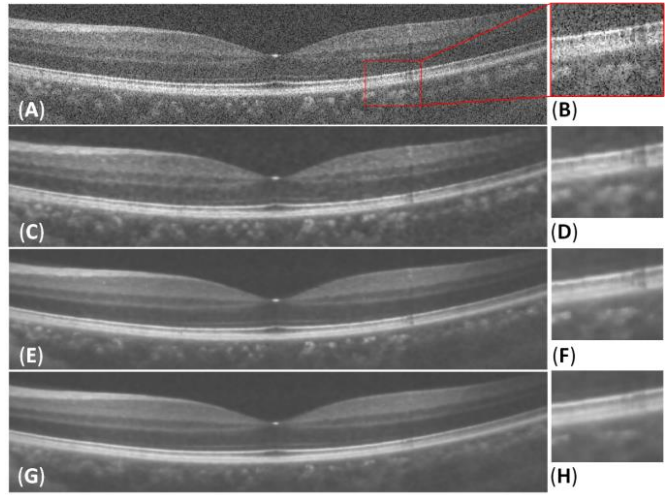

Fig. 4. The effect of different values of the parameter h for denoising a retinal OCT image. The second column shows a magnified region. Images (A) and (B) are the original Noisy Images, (C) and (D) are the MIFCN's outputs applying h = 1; (E) and (F) are the MIFCN's output applying h = 400; and (G) and (H) are the MIFCN's outputs applying h = 2000. This figure is better observed by zooming on a computer screen.

Quantitative metrics can be contributive in selecting an appropriate value for *h*. The mean values of some metrics which are obtained using different values of the parameter *h* are tabulated in Table IV. However, each metric has its own merits and drawbacks; therefore, visual inspection is still the most prominent tool.

Table IV

Mean of the MSR, CNR, ENL, and PSNR results for denoising 18 foveal images obtained using different values of the parameter h for the proposed MIFCN method

| *h* value | MSR | CNR | ENL | PSNR |
|---|---|---|---|---|
| 1 | 7.40 | 3.53 | 735.46 | 26.77 |
| 100 | 7.87 | 3.63 | 2557.01 | 27.14 |
| 200 | 8.12 | 3.69 | 2712.91 | 27.26 |
| 300 | 8.27 | 3.72 | 2744.02 | 27.33 |
| 400 | 8.38 | 3.75 | 2750.75 | 27.37 |
| **500** | 8.45 | 3.77 | **2753.79** | 27.39 |
| 600 | 8.51 | 3.79 | 2751.27 | 27.41 |
| 700 | 8.56 | 3.80 | 2748.64 | 27.42 |
| 800 | 8.59 | 3.82 | 2749.03 | 27.42 |
| 900 | 8.63 | 3.83 | 2748.38 | 27.43 |
| 1000 | 8.65 | 3.84 | 2747.28 | 27.43 |

For an ideal metric, it is natural to expect that an increase in the *h* constant value, would increase the metric value to a certain point, and then it begins to decrease. This is because increasing the *h* value leads to more contributions from other nearby images. This would reconstruct an image which is not faithful to the main input image. As observed in Table IV, the MSR and CNR metrics are not helpful in this experiment, because their values are on a continuous rise. The ENL and PSNR give more relevant results which are more consistent with the visual results in Fig. 4. For example, when the parameter $h$ is greater than 500 in Table IV, the ENL value begins to decrease, and the PSNR value remains almost constant. Figure 4 (F) and (H) demonstrate that when the $h$ constant value is much larger than 500, the small features become less visible, and the result becomes blurry. Therefore, for a given dataset of OCT images, both the quantitative metrics and the visual quality of the reconstruction results must be considered to find an appropriate value for $h$. Here, $h = 400$ is applied for all experiments in Sec. 5.5.

*5.7 EFFECTS OF THE NUMBER OF LAYERS*

In this section, it is revealed how the number of convolution layers could affect the performance of the proposed MIFCN method. In the architecture of MIFCN (Figs. 1 and 2), the main ingredients consist of $3 \times 3$ convolutions followed by LReLU activation layers, $1 \times 1$ convolution layers, and a pixel by pixel averaging module. Here, the focus is on varying the number of $3 \times 3$ convolution layers while keeping all other things unchanged.

The mean squared error (MSE) of five different configurations are tabulated in Table V. These configurations are indicated by MIFCN-A-B, where A is the number of $3 \times 3$ convolution layers for each branch (Fig. 2), and B is the number of $3 \times 3$ convolution layers right after the pixel by pixel averaging module (Fig. 1). A similar training set and training procedure are applied to learning the parameters of each configuration. Here, for evaluating each trained model, the learned model is used for reconstructing both the test and training sets. This is because changing the number of layers might easily lead to overfitting or underfitting. Therefore, evaluating errors for the training and test sets can provide more insights into the performance of a given model.

Table V

MSE for the training and test sets. The best test set MSE is shown in bold.

| Configuration | Training set | Test set |
| --- | --- | --- |
| MIFCN-3-0 | 84.64 | 120.55 |
| MIFCN-3-1 | 84.83 | **119.23** |
| MIFCN-3-2 | 85.64 | 120.28 |
| MIFCN-4-1 | 79.7 | 120.12 |
| MIFCN-2-1 | 94.18 | 123.13 |

In Table V, the configuration indicated by MIFCN-3-1 shows the main model, which is applied to all experiments in the previous sections. Removing the convolution layer after the pixel by pixel averaging module leads to a model (MIFCN-3-0) with slightly inferior test performance. The configurations MIFCN-3-2 and MIFCN-4-1 exhibit that adding more convolution layers cannot improve the performance of the main model and this is due to the lack of training data. The MIFCN-4-1 configuration clearly shows that the model needs more data since the MSE for the training set decreases significantly, but the test error does not improve. The last configuration in Table V shows that removing one convolution layer from each branch leads to a model (MIFCN-2-1) with limited modeling ability. For this model, the MSEs for the training and test sets are significantly higher than the other configurations. These quantitative comparisons reveal that the main model (MIFCN-3-1) offer a good trade-off between performance and complexity.

## 6. CONCLUSION

In this paper, a neural network (named MIFCN) is proposed for denoising SDOCT images. The proposed MIFCN method exploits a weighted averaging module inspired by the nonlocal mean method [19] to effectively capture useful information from nearby OCT images. We show how the parameters of the proposed MIFCN method can be learned in an end-to-end manner. Extensive experiments are run to compare the proposed MIFCN method with some of the well-known methods. The experimental results indicate the effectiveness of the proposed MIFCN method over the compared methods. It can effectively reduce noise while preserving the textures and layer structures. The proposed MIFCN method produces fewer artifacts compared to other methods. Therefore, it is not only useful for OCT image quality improvement, but also it might be a good preprocessing step for retinal layer segmentation methods. In the future, we would like to incorporate segmentation information into the proposed MIFCN method [14]. Also, we would like to extend the proposed MIFCN method for OCT image interpolation [14,15]. In addition, although we only considered the task of retinal OCT image denoising, the proposed MIFCN method might also be applied to denoising of other medical images.

## ACKNOWLEDGMENT

This work was supported by the National Natural Science Foundation of China under Grants 61501180 and 61771192.